\documentclass[preprint,12pt]{elsarticle}

\usepackage{amssymb}
\usepackage{amsmath}
\usepackage{multirow}
\usepackage{makecell}
\usepackage{graphicx}  

\journal{SPIE Medical Imaging: Image Processing}

\begin{document}

\begin{frontmatter}



\title{Divide to Conquer: A Field Decomposition Approach for Multi-Organ Whole-Body CT Image Registration}


\author[a]{Xuan Loc Pham} 
\author[a]{Mathias Prokop}
\author[a,b]{Bram van Ginneken}
\author[a]{Alessa Hering}

\affiliation[a]{organization={Department of Imaging, Radboudumc},
            city={Nijmegen},
            postcode={6525 GA}, 
            state={Gelderland},
            country={the Netherlands}}
\affiliation[b]{organization={Fraunhofer MEVIS},
            city={Bremen},
            country={Germany}}
\begin{abstract}
Image registration is an essential technique for the analysis of Computed Tomography (CT) images in clinical practice. However, existing methodologies are predominantly tailored to a specific organ of interest and often exhibit lower performance on other organs, thus limiting their generalizability and applicability. Multi-organ registration addresses these limitations, but the simultaneous alignment of multiple organs with diverse shapes, sizes and locations requires a highly complex deformation field with a multi-layer composition of individual deformations. This study introduces a novel field decomposition approach to address the high complexity of deformations in multi-organ whole-body CT image registration. The proposed method is trained and evaluated on a longitudinal dataset of 691 patients, each with two CT images obtained at distinct time points. These scans fully encompass the thoracic, abdominal, and pelvic regions. Two baseline registration methods are selected for this study: one based on optimization techniques and another based on deep learning. Experimental results demonstrate that the proposed approach outperforms baseline methods in handling complex deformations in multi-organ whole-body CT image registration.
\end{abstract}


\end{frontmatter}



\section{Introduction}
\label{sec:intro}

Image registration is a fundamental area of medical image analysis. Thanks to advancements in hardware development, deep learning-based image registration has increasingly shown significant potential in both research and clinical practice. However, current methods are primarily designed for single-organ registration, such as brain \cite{Junyu_2022}, lung \cite{Hering_2021}, or liver \cite{Pham_2024}, and often exhibit suboptimal performance when applied to other organs. This limitation necessitates a new dataset and retraining of models when changing the organ of interest, which significantly constrains their generalizability and broader applicability. Multi-organ registration addresses these challenges but introduces additional complexities. It requires the simultaneous optimization of multiple deformation fields rather than focusing on a single one. Additionally, the deformation fields for different organs can vary significantly in terms of size, shape (e.g., pancreas vs. liver), and location (e.g., pancreas vs. lung), which further complicates the registration process. 

\begin{figure} [ht]
\begin{center}
\begin{tabular}{c} 
\includegraphics[width=0.98\textwidth]{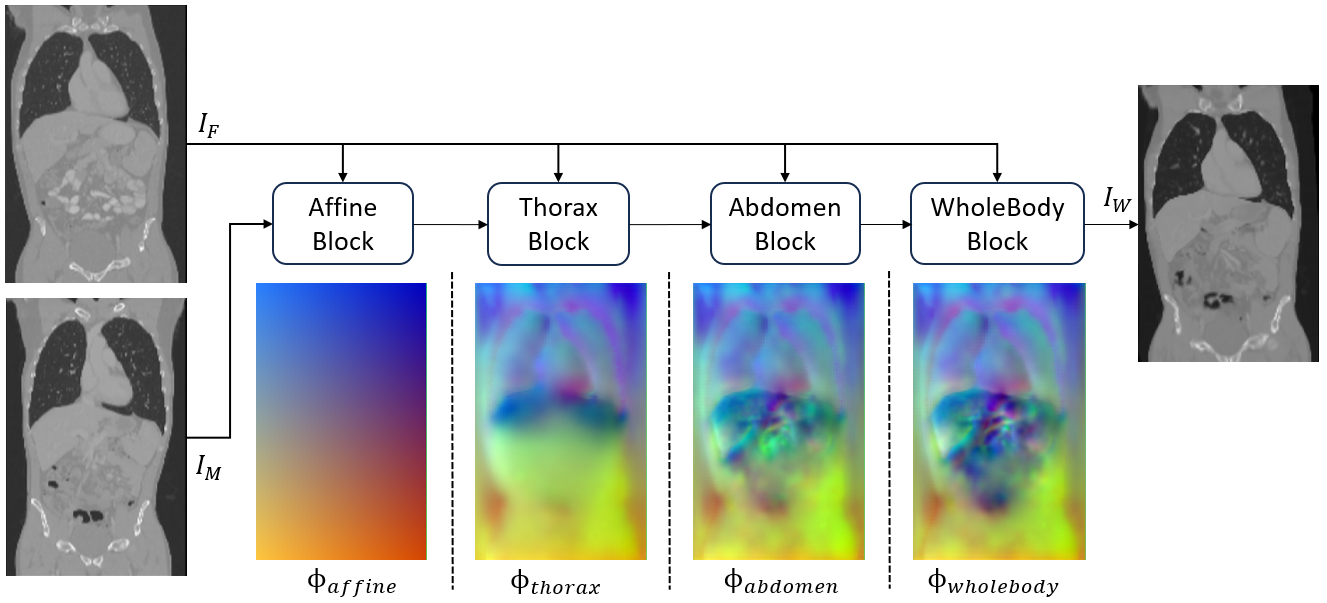}
\end{tabular}
\end{center}
\caption[example] 
{ \label{fig:method} 
Illustration of the \textit{Divide and Conquer} strategy. The proposed method comprises 4 registration blocks, each containing a UNet to generate the deformation field and a Spatial Transformer Network \cite{Jaderberg_2015} to generate the warped image. Each registration block is assigned a specific region of interest. Specifically, the \textit{Affine Block} checks the whole body volume for affine registration, the \textit{Thorax Block} and \textit{Abdomen Block} focus on the thorax and abdomen regions respectively, while the \textit{Wholebody Block} inspects all organs for refinement. Note that the cumulative field $\phi$ is the accumulation of all previous deformation fields $\Phi$ before it.
}
\end{figure}

This study introduces a novel displacement field decomposition approach to manage the high complexity of deformations in multi-organ whole-body CT image registration, as illustrated in Figure \ref{fig:method}. Our primary contributions are threefold. First, we propose dividing the complex deformation field into smaller, more manageable components based on specific tasks and regions. Specifically, instead of directing the model to analyze the entire body image simultaneously, which may lead to confusion, we assign each registration block to focus on a particular region and combine them using a multi-cascade strategy \cite{Pham_2024}. Additionally, a dedicated block is utilized for affine registration to mitigate the complexity of deformations before training deformable registration on specific regions. Secondly, we simultaneously utilize multiple segmentation labels of the organs of interest as supplementary information during the training phase, which enhances the precision in locating the organs. Finally, to the best of our knowledge, we are the first to develop a large-scale intra-patient CT dataset that encompasses the whole-body view, thereby contributing to advancing the field of multi-organ whole-body CT image registration. In this study, we choose four organs of interest: lung, liver, kidneys and pancreas, which considerably vary in size, shape and location in the body, to verify the effectiveness of the proposed method.

\section{METHOD} 

\subsection{Problem Formulation}
A volumetric registration task involves establishing the optimal alignment between two input images, as described by equations \eqref{IW}, \eqref{Q}. Here, \( I_F \), \( I_M \), and \( I_W \) represent the fixed, moving, and warped images within the spatial domain \(\Omega \subset \mathbb{R}^3\). The transformation field \(\Phi\) denotes the mapping between \( I_M \) and \( I_F \).
\begin{equation}
    I_W = I_M \circ \Phi \quad,
\label{IW}
\end{equation}
\begin{equation}
    \hat{\Phi}=\arg{\min_\Phi{\mathcal{L}\left(I_F,I_M,\Phi\right)}} \quad,
\label{Q}
\end{equation}
Existing image registration methods typically concentrate on optimizing $\Phi$ for a single organ of interest. However, in the context of multi-organ registration, $\Phi$ is extended as described in equation \eqref{O_compose}, which necessitates the parallel optimization of multiple deformation fields, thereby introducing a significant level of complexity.
\begin{equation}
    \Phi = \Phi_{lung} + \Phi_{liver} + \Phi_{kidney} + \Phi_{pancreas}
\label{O_compose}
\end{equation}

\subsection{Proposed Method}
In this study, we propose a task-based and region-based field decomposition approach, as seen in Equation \eqref{O_propose}, to cope with the high complexity of multi-organ registration. Specifically, the total deformation is explicitly divided into affine and non-rigid components. The non-rigid registration part is further divided into deformations in the thorax and abdomen regions. Finally, the field $\Phi_{wholebody}$ is added to refine and balance the impact between the thoracic and abdominal deformations, ensuring a cohesive and accurate alignment across the entire body.
\begin{equation}
    \Phi = \Phi_{affine} + \Phi_{thorax} + \Phi_{abdomen} + \Phi_{wholebody} 
\label{O_propose}
\end{equation}
Equation \eqref{L} represents the loss function used to optimize $\Phi$. The overall loss consists of three components: the similarity Mutual Information loss $\mathcal{L}_{MI}$, the segmentation overlapping Dice loss $\mathcal{L}_{DSC}$,
and the field regularization Bending Energy loss $\mathcal{L}_{BE}$. We also assign the weights $\alpha$, $\lambda$ and $\beta$ to each respective loss component to control their contributions to the overall loss function.  
\begin{equation}
\mathcal{L}\left(I_F,I_M,\Phi\right)=\alpha\mathcal{L}_{MI}\left(I_F,I_W\right)+\lambda\mathcal{L}_{DSC}\left(S_F,S_W\right)+\beta\mathcal{L}_{BE}\left(\Phi\right) 
\label{L}
\end{equation}

The proposed architecture is motivated from Equation \eqref{O_propose}, which comprises four blocks. Three deformable registration blocks use the basic UNet architecture with the encoder-decoder branch, while the affine block only needs the encoder branch to output 12 values for the affine transformation matrix. We first trained the affine block using pairs of intra-patient CT scans and their body segmentation labels. The resulting warped images and masks from the affine block then served as inputs to train each deformable registration block separately. During the training of the thorax block, the body segmentation mask and thoracic organ label (lung) were provided to better locate the organs of interest. Similarly, the body segmentation mask and abdominal organ labels (liver, kidneys and pancreas) were employed to train the abdomen block. Ultimately, we utilized the segmentation labels of all organs to train the whole-body registration block to refine and smooth out any conflicting regions between the thorax and abdomen blocks. All registration blocks were trained separately and unsupervisedly to optimize both time and hardware resources. The final deformation field is effectively obtained by summing the output displacements from all four blocks.

\section{RESULTS}
\subsection{Experimental Setup}
\subsubsection{Data Collection}
Experiments in this study were carried out based on an in-house longitudinal CT dataset consisting of 691 patients, each with two CT images taken at different time points. Most patients received therapy and surgery between the two imaging sessions. All scans in the dataset fully contain the whole body view, including the thorax, abdomen and pelvis regions. To ensure the variability and richness of the dataset, we selected patients from a broad age range and maintained a relatively balanced gender ratio. We also included scans obtained from various CT scanners in the hospital such as Siemens, Toshiba, Philips, GE and Canon. 

\subsubsection{Data Preprocessing}
In the preprocessing phase, we first adjusted the patient's position to the standard orientation and rescaled the intensity to ensure correct HU units if needed. Following this, we normalized the dataset and used \textit{TotalSegmentator} \cite{Wasserthal23} with its default weights to automatically generate segmentation labels for the organs of interest, including the lung, liver, kidneys, pancreas, and also the body. The body segmentation label was then used to crop the original images to remove any unnecessary information. Finally, we resampled the cropped images to a uniform dimension of 256x192 pixels with 160 slices.

\subsubsection{Training and Evaluation}
For the training phase, 591 pairs of intra-patient CT scans with segmentation labels were randomly selected from the dataset as inputs to the model. For the evaluation phase, the remaining 100 CT pairs with segmentation labels were visually reviewed by a PhD student with 2 years of experience in our group. Additionally, a sub-dataset of 100 inter-patient CT pairs was created by randomly pairing scans from the intra-patient data above. The proposed method was then evaluated on both intra-patient and inter-patient scenarios to comprehensively assess its ability to handle deformation fields of varying complexity. 

\subsection{Results and Discussions}
\renewcommand{\arraystretch}{2} 
\begin{table}[ht]
\caption{Evaluation results of intra-patient registration. The results are presented as mean ± standard deviation.} 
\vspace{10pt}  
\label{tab:intrareg}
\centering
\small
\resizebox{\linewidth}{!}{ 
\begin{tabular}{|p{2.6cm}|>{\centering\arraybackslash}p{1.6cm}|>{\centering\arraybackslash}p{1.6cm}|>{\centering\arraybackslash}p{1.6cm}|>{\centering\arraybackslash}p{1.6cm}|>{\centering\arraybackslash}p{1.6cm}|}
\hline
\textbf{Methods} & \textbf{Liver} & \textbf{Kidney} & \textbf{Pancreas} & \textbf{Lung} & \textbf{Folding (\%)} \\
\hline
Raw Data & \makecell{72.11 \\ ± 12.29} & \makecell{54.29 \\ ± 16.10} & \makecell{35.28 \\ ± 21.47} & \makecell{76.88 \\ ± 8.60} & 0 \\
\hline
VoxelMorph \cite{balakrishnan_2019} & \makecell{75.28 \\ ± 17.00} & \makecell{56.67 \\ ± 19.68} & \makecell{37.68 \\ ± 22.92} & \makecell{85.37 \\ ± 12.32} & \makecell{0.03 \\ ± 0.2} \\
\hline
Elastix \cite{klein_2010} & \makecell{91.90 \\ ± 4.70} & \makecell{84.28 \\ ± 9.88} & \makecell{63.11 \\ ± 20.20} & \makecell{\textbf{97.06} \\ ± 1.65} & \makecell{0.55 \\ ± 1.05} \\
\hline
Proposed method & \makecell{\textbf{93.22} \\ ± 5.01} & \makecell{\textbf{87.20} \\ ± 10.68} & \makecell{\textbf{63.49} \\ ± 20.88} & \makecell{96.70 \\ ± 2.78} & \makecell{0.98 \\ ± 0.54} \\
\hline
\end{tabular}
} 
\end{table}

Table \ref{tab:intrareg} presents the experimental results evaluating the registration performance on 100 pairs of intra-patient CT images. Overall, the proposed field decomposition approach outperforms other baseline methods across nearly all organs of interest. There is a considerable improvement in the Dice Similarity Coefficient (DSC) metric when comparing the Voxelmorph baseline with the proposed method (the statistical p-values of the t-test are less than 0.05 for all organs). When compared with Elastix, the proposed method still demonstrates noticeably better performance in the alignment of the liver or kidney regions (p-value $<$ 0.05). However, the difference is less obvious for the pancreas and lung organs (p-value $>$ 0.05). Regarding the folding metric (quantified as the percentage of negative values in the Jacobian determinant of the deformation field), the field decomposition technique introduces more folding regions than baseline methods. Nevertheless, the distortion remains generally tolerable with less than 1$\%$ of the whole image volume. 

\begin{table}[ht]
\caption{Evaluation results of inter-patient registration. The results are presented as mean ± standard deviation.} 
\vspace{10pt}  
\label{tab:interreg}
\centering
\small
\resizebox{\linewidth}{!}{ 
\begin{tabular}{|p{2.6cm}|>{\centering\arraybackslash}p{1.6cm}|>{\centering\arraybackslash}p{1.6cm}|>{\centering\arraybackslash}p{1.6cm}|>{\centering\arraybackslash}p{1.6cm}|>{\centering\arraybackslash}p{1.6cm}|}
\hline
\textbf{Methods} & \textbf{Liver} & \textbf{Kidney} & \textbf{Pancreas} & \textbf{Lung} & \textbf{Folding (\%)} \\
\hline
Raw Data & \makecell{49.37 \\ ± 18.53} & \makecell{25.49 \\ ± 12.48} & \makecell{9.40 \\ ± 10.68} & \makecell{64.16 \\ ± 11.15} & 0 \\
\hline
Elastix \cite{klein_2010} & \makecell{71.14 \\ ± 17.62} & \makecell{40.80 \\ ± 17.83} & \makecell{19.94 \\ ± 14.31} & \makecell{90.51 \\ ± 15.09} & \makecell{5.62 \\ ± 4.32} \\
\hline
Proposed method & \makecell{\textbf{78.17} \\ ± 17.82} & \makecell{\textbf{52.81} \\ ± 21.04} & \makecell{\textbf{23.40} \\ ± 17.36} & \makecell{\textbf{93.40} \\ ± 6.14} & \makecell{\textbf{2.41} \\ ± 1.13} \\
\hline
\end{tabular}
} 
\end{table}

\begin{figure} [t]
\begin{center}
\begin{tabular}{c} 
\includegraphics[width=0.98\textwidth]{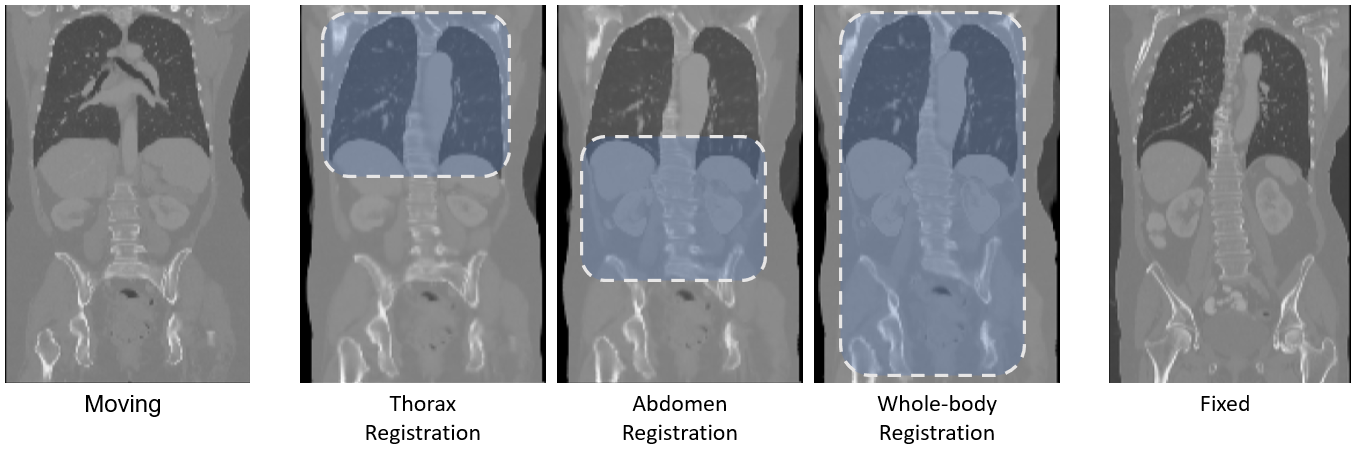}
\end{tabular}
\end{center}
\caption[experiment] 
{ \label{fig:experiment} 
Illustration of an inter-patient multi-organ registration case with large deformations. Images 2, 3 and 4 show warped images after thorax registration, abdomen registration and whole-body registration, respectively. The blue region indicates the region of focus for each registration block.
}
\end{figure} 

Table \ref{tab:interreg} and Figure \ref{fig:experiment} show the experimental results of the ablation study on inter-patient registration to better understand the performance of the proposed method in scenarios involving very large deformations. In general, the improvement of the proposed method over Elastix is more significant in this experiment. Specifically, there is a 12$\%$ difference in DSC for the kidney region and 7$\%$ for the liver region. Additionally, the field decomposition technique also results in 3.21$\%$ fewer folding regions than the traditional registration algorithm.

In summary, the field decomposition technique can effectively handle the parallel optimization of multiple deformation fields compared to both traditional and deep learning-based registration baseline methods. Additionally, experiment results show that incorporating segmentation label guidance during the training phase also contributes largely to these promising results. The proposed approach also demonstrates improved robustness and stability, as shown in the ablation study involving very large deformations. In this experiment, the proposed method increases folding by a factor of 2.4, while Elastix requires a tenfold increase to handle inter-patient deformations. While the proposed method generally performs well on large and medium-sized organs, its performance on smaller organs, such as the pancreas, still needs further improvement. In future studies, we plan to expand the current dataset, involve radiologists and medical experts, and include segmentation masks for more organs from various body regions to improve the quality of the current cohort. In addition, we will also experiment on a wide range of external datasets to achieve a more comprehensive evaluation of the proposed method.  

\section{CONCLUSIONS}
\label{sec:conclusions}

This study introduces a field decomposition approach for the simultaneous registration of multiple organs in the body. The proposed method explicitly divides the complex registration task into four smaller displacement fields based on their tasks or regions of focus. Four blocks are separately trained with segmentation label guidance to generate the displacement fields for the thorax region, abdomen region, whole body region and also for the affine transformation. For this research, we have built a large-scale intra-patient whole-body CT dataset of 691 pairs. Experimental results demonstrate that explicitly assigning a registration block to learn the displacement in each specific region effectively handles the highly sophisticated deformation field in multi-organ registration, rather than processing the entire image at once. These promising results in multi-organ registration are expected to enhance the applicability of deep learning-based image registration, thereby contributing to bridging the gap between research and clinical practice. 




\bibliography{ref} 
\bibliographystyle{elsarticle-num} 




\end{document}